\documentclass{article}

\usepackage{arxiv}

\usepackage[utf8]{inputenc} 
\usepackage[T1]{fontenc}    
\usepackage{hyperref}       
\usepackage{url}            
\usepackage{booktabs}       
\usepackage{amsfonts}       
\usepackage{nicefrac}       
\usepackage{microtype}      
\usepackage{lipsum}

\usepackage{microtype}
\usepackage{graphicx}
\usepackage{subfigure}
\usepackage{booktabs} 
\usepackage{paralist}

\usepackage{amsmath}
\usepackage{amssymb}

\newenvironment{tight_enumerate}{
\begin{enumerate}
  \setlength{\itemsep}{0pt}
  \setlength{\parskip}{0pt}
}{\end{enumerate}}

\title{RecNets: Channel-wise Recurrent Convolutional Neural Networks}

\author{
  George Retsinas\\
  School of ECE, NTUA, Athens, Greece\\
  \texttt{gretsinas@central.ntua.gr} \\
   \And
 Athena Elafrou\\
  School of ECE, NTUA, Athens, Greece\\
  \texttt{athena@cslab.ece.ntua.gr} \\
  \And
 Georgios Goumas\\
  School of ECE, NTUA, Athens, Greece\\
  \texttt{goumas@cslab.ece.ntua.gr} \\
  \And
 Petros Maragos\\
  School of ECE, NTUA, Athens, Greece\\
  \texttt{maragos@cs.ntua.gr} \\
}

\begin{document}

\maketitle

\begin{abstract}
In this paper, we introduce Channel-wise recurrent convolutional neural networks (RecNets), a family of novel, compact neural network architectures for computer vision tasks inspired by recurrent neural networks (RNNs). 
RecNets build upon Channel-wise recurrent convolutional (CRC) layers, a novel type of convolutional layer that splits the input channels into disjoint segments and processes them in a recurrent fashion.
In this way, we simulate wide, yet compact models, since the number of parameters is vastly reduced via the parameter sharing of the RNN formulation.
Experimental results on the CIFAR-10 and CIFAR-100 image classification tasks demonstrate the superior size-accuracy trade-off of RecNets compared to other compact state-of-the-art architectures.
\end{abstract}

\section{Introduction}
\label{sec:intro}

Convolutional neural networks (CNNs) deliver state-of-the-art results in a variety of computer vision tasks, including image classification~\cite{NIPS2012_4824}, image segmentation~\cite{Long_2015_CVPR} and object recognition~\cite{Ren:2017:FRT:3101720.3101780}. The general trend has been to design deeper and more sophisticated network architectures in order to build higher accuracy models~\cite{Szegedy_2015_CVPR,NIPS2015_5850,He_2016_CVPR}. However, the increased memory and computational requirements of such models pose serious challenges to the system designer and hinder their deployment on resource-constrained devices. This has lead researchers to explore various approaches for improving the efficiency of CNNs. Such approaches can be generally categorized into either designing and training compact, yet accurate architectures from scratch~\cite{NIPS2015_5647,Szegedy_2015_CVPR,Howard2017MobileNetsEC,8252668} or compressing pre-trained models via pruning~\cite{8237417,8237803,NIPS2016_6504}, quantization~\cite{NIPS2016_6573,Jacob_2018_CVPR}, decomposition~\cite{NIPS2013_5025,NIPS2014_5544,Liu_2015_CVPR} or distillation~\cite{polino2018model}, with the aim of preserving model accuracy.

In this work we present~\emph{RecNets}, a family of neural network architectures for computer vision tasks inspired by Recurrent Neural Networks (RNNs) ~\cite{lipton2015critical}. Driven by the need for compact, yet accurate models, RecNets incorporate two important features that allow them to achieve this goal: (a) parameter sharing through a recurrent formulation of convolutional layers and (b) wide feature maps---which have been shown to improve network performance~\cite{Zagoruyko2016WideRN}. The recurrent logic embedded in the design of RecNets can be viewed as a different approach to the concept of cross-layer connectivity paths established by recent architectures such as DenseNets~\cite{Huang_2017_CVPR} and DualPathNets~\cite{NIPS2017_7033}.

The main building block of RecNets comprises a \emph{Channel-wise Recurrent Convolutional} (CRC) layer followed by a pointwise convolutional layer. 
A CRC layer differs from a typical convolutional layer in that it splits the input channels into disjoint segments and processes them in a recurrent fashion. 
In this way, a CRC layer reuses the same parameters for every segment, vastly reducing the number of parameters required to process the input. 
At the same time, the hidden state of the layer, generated by the recurrent process, captures information along the sequence of input channels, therefore enabling information flow between successive segments. 
The output of the CRC layer is the concatenation of the $d$ generated hidden variables, which can be considered as a sequence generated by the underlying RNN.  
Therefore, CRC can approximate wide layers in a way similar to grouped convolutions~\cite{NIPS2012_4824}, but retains a group connectivity through the recurrent formulation. 
The reduced representational power due to parameter sharing is replenished to some degree by the hidden state of the recurrent formulation. 


We evaluate RecNets on two popular benchmark datasets (CIFAR-10, CIFAR-100) considering architectures of at most a few million parameters ($<10M$). 
Our models achieve a notable size-accuracy trade-off, outperforming state-of-the-art models. For instance, compared to the popular MobileNet architecture~\cite{Howard2017MobileNetsEC} which has an accuracy of $73.65\%$ on CIFAR-100 at 3.3M parameters, a RecNet with approximately the same number of parameters achieves a higher accuracy of $79.01\%$, while a much smaller RecNet (0.3M parameters) achieves a similar accuracy of $73.68\%$.

\section{RecNet Architecture}
\label{sec:recnet}

In this section we first introduce the backbone of the RecNet architecture, code-named Channel-wise Recurrent Convolutional (CRC) layer. Along with the CRC layer definition, we consider several design choices of the layer and report possible advantages and drawbacks. We then introduce a family of end-to-end architectures, called RecNets, that will be evaluated in this work along with their hyper-parameters. 

\subsection{Channel-wise Recurrent Convolutional Layers}
\label{sec:recnet:CRC}

In~\cite{NIPS2017_7033} the authors hint that residual networks~\cite{He_2016_CVPR} and their variations have a formulation similar to RNNs regarding the usage of successive layers. Driven by this observation, the main idea behind the RecNet architecture is to explicitly embed this recurrent logic in the network structure by replacing a typical convolutional layer by multiple successive smaller layers that form an RNN across the channel dimension. In short, we propose a recurrent formulation of a convolutional layer by splitting its input and output feature maps across the layer's channels into segments; the input segments form a sequence that is fed to an RNN, while the output sequence, generated by the RNN, consists of the output segments. We will henceforth refer to such layers as \emph{Channel-wise Recurrent Convolutional} (CRC) layers.

\subsubsection{Layer Definition}
\label{sec:recnet:CRC:definition}

Let $\mathbf{x}\in \mathbb{R}^{C_{in}\times H\times W}$ be the input tensor and $\mathbf{h}\in \mathbb{R}^{C_{out}\times H\times W}$ be the output tensor of a CRC layer, where $C_{in}$, $C_{out}$ are the number of input and output channels respectively and $W$, $H$ the spatial dimensions of each individual channel. We split $\mathbf{x}$ and $\mathbf{h}$ into $d$ segments across the channel dimension, i.e.\ $\mathbf{x} = (\mathbf{x}_0, \mathbf{x}_1, \dots, \mathbf{x}_{d-1})$ and $\mathbf{h} = (\mathbf{h}_0, \mathbf{h}_1, \dots, \mathbf{h}_{d-1})$ so that $\mathbf{x}_i\in \mathbb{R}^{S_{in}\times H\times W}$ and $\mathbf{h}_i\in \mathbb{R}^{S_{out}\times H\times W}$ for $i\in [0,d-1]$, where $S_{in}=C_{in}/d$ and $S_{out}=C_{out}/d$ are the input and output channels of each segment respectively. Assuming the $\{\mathbf{x}_i\}$ segments form the input sequence of an RNN and the $\{\mathbf{h}_i\}$ segments comprise its hidden state, we get the following recurrent formulation:
\begin{equation}
\mathbf{h}_i = \sigma(f_x(\mathbf{x}_i) + f_h(\mathbf{h}_{i-1})),
\label{eq1}
\end{equation}
\noindent where $\sigma()$ is a nonlinear activation function and $f_h()$, $f_x()$ are the transformation functions for the hidden state and input respectively. The output feature map $y$ of the CRC layer is formed by the concatenation of the hidden state segments $\{\mathbf{h}_i\}$. 

Let $\mathbf{W_h}\in \mathbb{R}^{S_{out}\times S_{out}\times k\times k}$ be the weight tensor for the hidden state, $\mathbf{W_x}\in \mathbb{R}^{S_{out}\times S_{in}\times k\times k}$ be the weight tensor for the input and $\mathbf{b}\in \mathbb{R}^{S_{out}}$ be the bias. Contrary to typical RNNs, we define every transformation of the input and hidden state as a convolutional layer with $k\times k$ filters so that Eq.~(\ref{eq1}) becomes:
\begin{equation}
\mathbf{h}_i = 
\begin{cases}
\sigma(\mathbf{x}_i \circledast \mathbf{W_x} + \mathbf{b}), i = 0\\
\sigma(\mathbf{x_i} \circledast \mathbf{W_x} + \mathbf{h}_{i-1} \circledast \mathbf{W_h} + \mathbf{b}), \,\forall i \in [1, d-1].
\end{cases}
\label{eq2}
\end{equation}
In this way the CRC layer simulates a typical convolutional layer with $C_{in} \cdot C_{out} \cdot k \cdot k$ parameters using only $(S_{in} + S_{out}) \cdot S_{out} \cdot k \cdot k = (C_{in} + C_{out}) \cdot C_{out} \cdot k \cdot k\cdot \frac{1}{d^2}$ parameters. The term $\frac{1}{d^2}$ can significantly reduce the layer's parameters when using an appropriate number of steps (e.g $d=10$). To give an example, a CRC layer with 160 input and 640 output channels both split into 10 segments and has a total of 47,360 parameters (compared to 921,600 parameters for a typical $3\times 3$ convolutional layer with the same input and output channels). This reduction is the result of re-using the same weights across each step $i=0,\dots,d-1$ and can lead to very compact architectures. The receptive field of the CRC layer depends on the number of recurrent steps $d$, therefore its representational power is tightly coupled to this hyper-parameter. By default, we use $3\times 3$ convolution filters for processing both the input and hidden state of each step of the CRC layer. However, we will also explore the usage of $1 \times 1$ filters for either $\mathbf{W_h}$ or $\mathbf{W_x}$ in Section~\ref{sec:exp:crc:kernel} to further reduce the number of parameters.

If we omit the $\mathbf{W_h}$ term from Eq.~(\ref{eq2}), which corresponds to the ``history" of already processed feature maps, the CRC layer degenerates to a grouped convolution with shared parameters across the groups, i.e. $\mathbf{h_i} = \sigma(\mathbf{x_i} \circledast \mathbf{W_x}), \, i \in [0, d-1]$. However, the history term $\mathbf{h_{i-1}} \circledast \mathbf{W_h}$ carries significant representational power, since it combines and encodes several previous feature maps, something that has been shown to strengthen feature propagation and encourage feature reuse~\cite{Huang_2017_CVPR}.

The computational cost of a CRC layer in terms of FLOPs is
$H \cdot W \cdot (C_{in} + C_{out}) \cdot C_{out} \cdot k \cdot k\cdot \frac{1}{d} \,\text{(convolutions)} + 2 \cdot H \cdot W \cdot C_{out} \,\text{(additions)}$, which is practically equivalent to applying two successive grouped convolutions of size $C_{in} \times C_{out} \times k \times k$ and $C_{out} \times C_{out} \times k \times k$ respectively, both using $d$ groups. 
Even though the parameters are reduced by a quadratic term ($1/d^2$) and the computational cost is linearly reduced ($1/d$), the drawback of the proposed layer compared to typical (grouped) convolutional layers lies on the fact that the sequential nature of the recurrent formulation cannot be effectively parallelized.  

In short, a CRC layer, which we will henceforth denote as CRC$(S_{in}$, $S_{out}$, $d)$, is defined in terms of the following hyper-parameters: 
1) $S_{in}$: number of input channels per segment, 2) $S_{out}$: number of output channels per segment and 
3) $d$: number of segments (equivalent to layer's depth).



\subsubsection{Non-Linearity}
\label{sec:recnet:CRC:nonlin}

Another differentiation from the typical RNN formulation involves the non-linear function $\sigma()$ in Eq.~(\ref{eq2}). We distinguish three cases of interest that will be evaluated in Section~\ref{sec:exp:crc:nonlin}: 
\begin{tight_enumerate}

\item \emph{ReLU non-linearity}: The simple case of using the ReLU function, which is the most popular choice of non-linearity in CNNs. 
\item \emph{BN + ReLU non-linearity}: Use of a Batch Normalization (BN)~\cite{Ioffe:2015:BNA:3045118.3045167} layer along with the ReLU function. To add a degree of freedom, the BN layer is not shared across the $d$ steps, but instead we use $d$ separate BN layers and drop the bias matrix $b$ in Eq.~(\ref{eq2}). We denote each BN layer along with the ReLU non-linearity as $\sigma_i()$. This choice barely affects the number of parameters since the BN layer parameters are linear to the number of channels. Equation~(\ref{eq2}) now becomes:
\begin{equation}
\mathbf{h}_i = 
\begin{cases}
\sigma_i(\mathbf{x_i} \circledast \mathbf{W_x}), i = 0\\
\sigma_i(\mathbf{x_i} \circledast \mathbf{W_x} + \mathbf{h}_{i-1} \circledast \mathbf{W_h}), \,\forall i \in [1, d-1].
\end{cases}
\label{eq3}
\end{equation}
\item \emph{Linear recursion}:
An interesting case is the adoption of a linear recurrent formulation by dropping the $\sigma()$ term:
\begin{equation}
\mathbf{h}_i = 
\begin{cases}
\mathbf{x_i} \circledast \mathbf{W_x}, i = 0\\
\mathbf{x_i} \circledast \mathbf{W_x} + \mathbf{h}_{i-1} \circledast \mathbf{W_h} + \mathbf{b}, \,\forall i \in [1, d-1].
\end{cases}
\label{eq4}
\end{equation}
An important property of a linear recurrent formulation is that the output can be re-written without the intermediate hidden variables, as shown in Eq.~(\ref{eq5}), and thus the layer can be parallelized at the cost of pre-computing the convolved tensors of Eq.~(\ref{eq5}). 
\begin{equation}
\mathbf{h}_i = \sum_{j=0}^{i} \mathbf{x}_j \circledast (\mathbf{W_x} \circledast \mathbf{W_h}^{\circledast (i-j)} ) + \sum_{j=0}^{i} \mathbf{b} \circledast (\mathbf{W_h}^{\circledast j} )
\label{eq5}
\end{equation}
Contrary to the previous cases where the non-linearity is applied at each recursion step, in this case we place a BN layer along with a ReLU non-linearity at the output of the CRC layer.
\end{tight_enumerate}

\subsection{Overall Architecture}
\label{sec:recnet:arch}

\begin{figure}
    \centering
    \includegraphics[width=.65\textwidth]{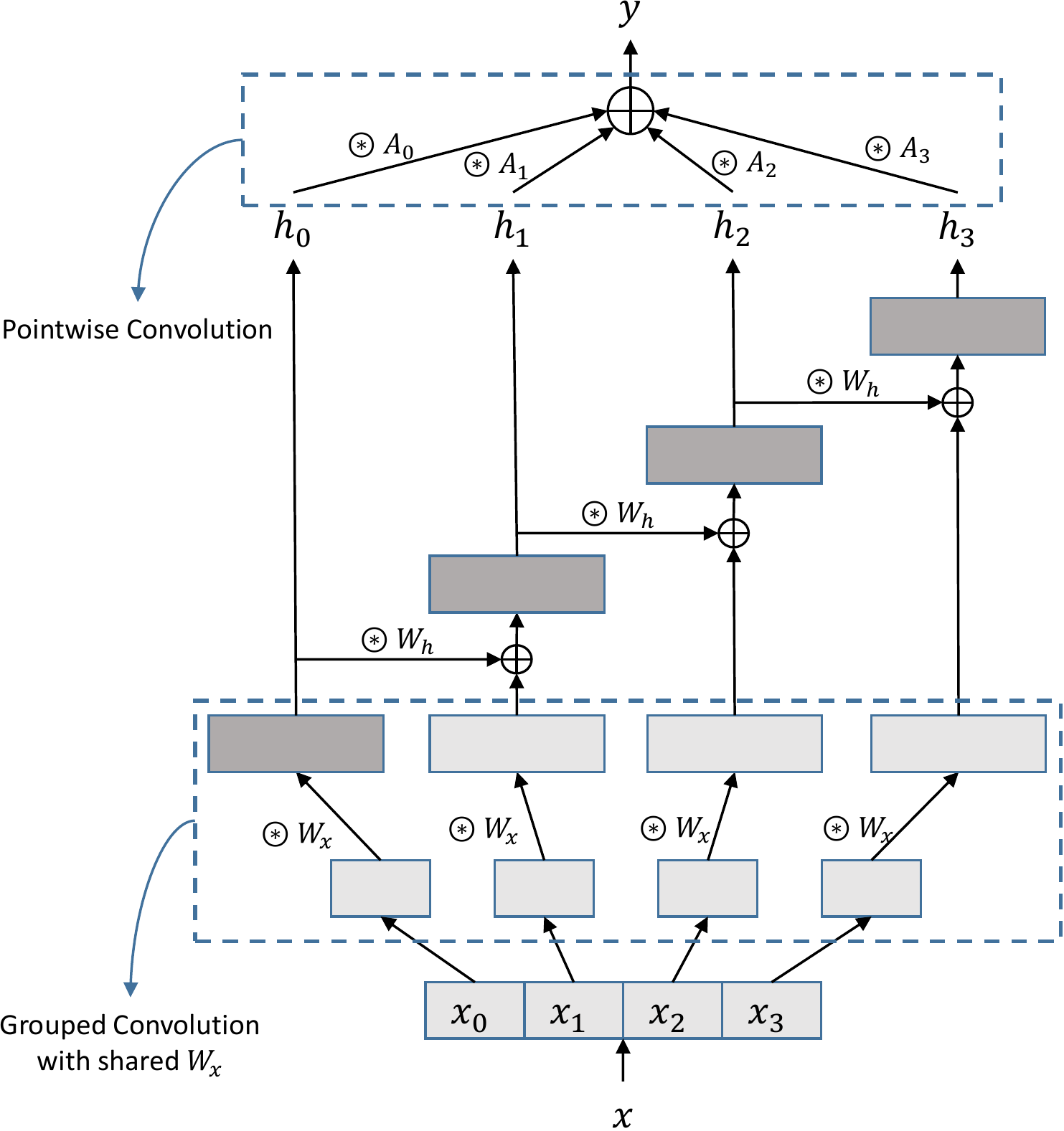}
    \caption{Visualization of a Recurrent module with $d=4$ segments. Feature map $\mathbf{x}$ is segmented into $\{\mathbf{x_i}\}$, forming a sequence of length $d$. Each segment has $S_{in}$ channels and is transformed into a new feature map of $S_{out}$ channels. This procedure is equivalent to a grouped convolution with shared weights ($\mathbf{W_x}$). The new sequence $\{\mathbf{h_i}\}$ is generated using the recurrent weight $\mathbf{W_h}$ and corresponds to the output of the CRC layer. The Recurrent module output $\mathbf{y}$ consists of $C_{out}$ channels and can be computed by a linear transformation, i.e. a point-wise convolution, over the CRC output $\{\mathbf{h_i}\}$.}
    \label{fig:rec}
\end{figure}

Using the proposed CRC layer as the backbone, we design a family of neural network architectures, called RecNets, by using successive CRC layers connected by $1 \times 1$ convolutions. Specifically, each CRC layer is followed by a \emph{Transition Block} layer, denoted TB($C_{in}$, $C_{out}$), of $C_{in}$ input channels and $C_{out}$ output channels which consists of a $1 \times 1$ convolutional layer followed by a BN layer and a ReLU non-linearity. In essence, the role of a TB layer is to perplex the channels that are generated by the different segments in a CRC layer so that each input segment of the following CRC layer carries information about the entire output of the previous CRC layer. 
Using CRCs alone, whilst they perform complex computations of high receptive field, they have an inherently imbalanced input-output segment correlation (e.g. $\mathbf{h}_0$ has seen only $\mathbf{x}_0$, while $\mathbf{h}_{d-1}$ has seen the entire $\mathbf{x}$ - see Eq.~(\ref{eq5})). 
Thus, if we create a network consisting only of CRC layers, this structural imbalance would be propagated and the leftmost segment, at any depth, would always be very narrow-sighted (similar concept to DenseNets~\cite{Huang_2017_CVPR}).
We will henceforth refer to the composition of a CRC layer with a TB layer as a \emph{Recurrent} module. Figure~\ref{fig:rec} provides an overview of a Recurrent module with $d=4$ segments.

\begin{table}[t]
\begin{center}
\begin{tabular}{lcc}
\toprule
 & Output & Output \\
Layer/Block & Channels & Size \\
\midrule\midrule
CONV ($3\times3$) + BN + ReLU & $S_1\cdot d_{1}$ & $32\times 32$ \\
\midrule
CRC ($S_1$, $e\cdot S_1$, $d_{1}$) & $e\cdot S_1\cdot d_{1}$ & $32\times 32$\\
TB ($e\cdot S_1\cdot d_{1}$, $S_1\cdot d_{1}$) & $S_1\cdot d_{1}$ & $32\times 32$\\
CRC ($S_1$, $e\cdot S_1$, $d_{1}$) & $e\cdot S_1\cdot d_{1}$  & $32\times 32$\\
TB($e\cdot S_1\cdot d_{1}$, $S_2\cdot d_{2}$) & $S_2\cdot d_{2}$ & $32\times 32$\\
\midrule
Max Pooling ($2\times 2$) &  $S_2\cdot d_{2}$ & $16\times 16$\\
\midrule
CRC ($S_2$, $e\cdot S_2$, $d_{2}$) & $e\cdot S_2\cdot d_{2}$ & $16\times 16$\\
TB($e\cdot S_2\cdot d_{2}$, $S_2\cdot d_{2}$) & $S_2\cdot d_{2}$ & $16\times 16$\\
CRC ($S_2$, $e\cdot S_2$, $d_{2}$) & $e\cdot S_2\cdot d_{2}$ & $16\times 16$\\
TB ($e\cdot S_2 \cdot d_{2}$, $S_3 \cdot d_{3}$) & $S_3 \cdot d_{3}$ & $16\times 16$\\
\midrule
Max Pooling ($2\times 2$) & $S_3 \cdot d_{3}$ & $8\times 8$\\
\midrule
CRC ($S_3$, $e\cdot S_3$, $d_{3}$) & $e\cdot S_3\cdot d_{3}$ & $8\times 8$\\
TB ($e\cdot S_3\cdot d_{3}$, $i_{2} \cdot d_{3}$) & $S_3\cdot d_{3}$ & $8\times 8$\\
CRC ($S_3$, $e\cdot S_3$, $d_{3}$) & $e\cdot S_3\cdot d_{3}$ & $8\times 8$\\
TB ($e\cdot S_3 \cdot d_{3}$, $S_3\cdot d_{3}$) & $S_3 \cdot d_{3}$ & $8\times 8$\\
\midrule
Average Pooling ($8\times 8$) & $S_3\cdot d_{3}$ & $1\times 1$\\
\midrule
Linear ($S_3\cdot d_{3}$, $n_{classes}$) & $n_{classes}$ & $1\times 1$\\
\bottomrule
\end{tabular}
\end{center}
\caption{RecNet($e$, $S_1$, $S_2$, $S_3$, $d_1$, $d_2$, $d_3$) architecture for CIFAR.}
\label{table:arch}
\end{table}


In the context of the RecNet architecture, a CRC layer expands the channels' dimension, while the subsequent TB layer shrinks it, preparing the input for the next CRC layer. 
If we fix the channel expansion at each CRC layer using an expansion parameter $e$ ($S_{out} = e * S_{in}$ for every CRC layer), we can describe the whole network through $e$ along with the input segment channels $S_{in}$ and the depth $d$ of each CRC layer.
In this work, we experiment with a RecNet network composed of three pairs of Recurrent modules, which we will henceforth denote as \emph{RecNet($e$, $S_1$, $S_2$, $S_3$, $d_1$, $d_2$, $d_3$)}.
The $S_i$ hyper-parameters correspond to the segments' input channels $S_{in}$ of the $i$-th pair of Recurrent modules.
For short, we also use the acronym \emph{RecNet-$w$-$d$}, where $w$ is the maximum width (\#channels) of the model ($w = e * \max\{S_i\cdot d_i\}$) and $d$ is the overall depth of the CRC layers ($d = \sum{d_i}$).  
The complete RecNet architecture is given in Table~\ref{table:arch}(b).

\subsection{Computational Simplification}
\label{sec:recnet:simpl}

The expansion parameter $e$ (see Section~\ref{sec:recnet:arch}) allows us to simulate wide layers, which may have great representational power (see Wide ResNets~\cite{Zagoruyko2016WideRN}). This, however, comes at the cost of large intermediate results between CRC and TB layers ($e\cdot C_{in} \times H \times W$), which increase memory requirements during inference. In practice, the CRC layer creates wide feature maps, while the TB layer reduces them in order to generate the input for the next CRC layer.
Nevertheless, the recurrent formulation enables us to rewrite the successive use of CRC and TB layers in a more compact way and merge them into a single layer.


Assume that a Recurrent module is denoted as Rec($S_{in}$, $S_{out}$, $C_{out}$, $d$) = \{CRC($S_{in}$, $S_{out}$, $d$), TB($d\cdot S_{out}$, $C_{out}$)\}. For simplicity, let the TB layer consist only of a $1 \times 1$ convolutional layer, whose weight matrix is $\mathbf{A}$. Since the input channels dimension of the TB layer is $d \cdot S_{out}$, the matrix $\mathbf{A}$ is sized $C_{out} \times d\cdot S_{out}$ and can be divided into $d$ groups as follows: $\mathbf{A} = [\mathbf{A}_0, \dots, \mathbf{A}_{d-1}]$, $\mathbf{A}_i : C_{out} \times S_{out}$.
If $\mathbf{h}^p = [\mathbf{h}_0^p, \dots, \mathbf{h}_{d-1}^p]^\intercal$ are the output segments of CRC for a specific pixel $p$, then the output of the subsequent TB $y^p$ for the pixel $p$ can be expressed as:
\begin{align}
\mathbf{y}^p = \mathbf{A}\cdot \mathbf{h}^p = \begin{bmatrix} \mathbf{A}_0  \dots  \mathbf{A}_{d-1} \end{bmatrix} \begin{bmatrix} \mathbf{h}_0^p \\ \dots \\ \mathbf{h}_{d-1}^p \end{bmatrix} = \sum_{0}^{d-1}{\mathbf{A}_i\cdot \mathbf{h}_i^p}
\end{align}

Based on the above formulation, the output of a Recurrent module can be computed as follows:
\begin{align}
\label{eq:simpl1}
\mathbf{y} = \sum_{0}^{d-1}{\mathbf{h}_i \circledast \mathbf{A_i}}\, , \quad
\mathbf{h}_i = 
\begin{cases}
\sigma(\mathbf{x}_i \circledast \mathbf{W_x} + \mathbf{b}), i = 0\\
\sigma(\mathbf{x_i} \circledast \mathbf{W_x} + \mathbf{h}_{i-1} \circledast \mathbf{W_h} + \mathbf{b}), \,\forall i \in [1, d-1],
\end{cases}
\end{align}

Using the above notation, we obtain the same results without the concatenation of the output sequence $\{ \mathbf{h_i} \}$ of the CRC layer, which results in large intermediate representations. Instead we use a summation over the separate responses of $\{ \mathbf{A_i} \}$ at each step of the RNN formulation using an intermediate feature map of dimension $S_{out}$. This approach is depicted at the top of Figure~\ref{fig:rec}.

\section{Experimental Results}

In this section we first evaluate different options related to the design of the CRC layer (see Section~\ref{sec:recnet:CRC}). We then explore RecNet's hyper-parameters controlling the depth and the width of the network, and, finally, we compare the proposed architecture against state-of-the-art networks. 

The experiments are performed on the popular CIFAR datasets~\cite{krizhevsky2009learning}.
The CIFAR-10 and CIFAR-100 datasets consist of $32\times32$ color images, corresponding to 10 and 100 classes respectively. Both datasets are split into 50,000 train and 10,000 test images. We employ data augmentation during training following the de facto standard methodology, consisting of horizontal flips and random crops. A simple mean/std normalization (per channel) is used on the input images. 
Training on CIFAR is performed using the SGD algorithm with Nesterov's momentum with the initial learning rate set to 0.1, weight decay to 0.0005, dampening to 0, momentum to 0.9 and minibatch size to 64. The overall epochs are set to 200, while the learning rate is changing according to a cosine annealing schedule \cite{loshchilov2016sgdr}. The learning rate scheduler is restarted at 20, 60 and 120 epochs.

\subsection{Exploration of DCR layer variants}
\label{sec:exp:CRC}

\subsubsection{Non-linearity}
\label{sec:exp:crc:nonlin}

As we presented earlier (see Section~\ref{sec:recnet:CRC:nonlin}), one differentiation in the design of the CRC layer from the typical RNN formulation is the choice of the non-linearity function $\sigma()$. In order to evaluate the importance of this choice, we evaluate four different cases: 
1) ReLU (typical RNN approach),
2) shared BN layer for all segments along with ReLU,
3) separate BN layers for each segment along with ReLU and
4) no $\sigma()$ function at all, i.e. evaluate a linear recursion formulation.
For the evaluation we used the RecNet(4,8,16,32,10,10,10) architecture and the results are summarized at Table \ref{table:CRC}(a). The shared BN approach performs poorly, since it significantly constrains the feature map value range. 
The ReLU non-linearity, which corresponds to the usual RNN usage, performs considerably well, but is on par with the computationally simpler linear recursion approach. Note that the non-linear recursion and the linear recursion have very similar accuracy even though we expected the non-linearity to play a crucial role in high-performing recursive layers. Nevertheless, using separate BN layers at each segment along with the ReLU function leads to non-trivial improvement in accuracy at the cost of a minimal increase of the network's parameters. 
Therefore, the separate BN approach will be used for the rest of the experimental section.

\vspace{-8pt}
\subsubsection{Expansion Hyper-Parameter}
The expansion hyper-parameter $e$ controls the number of output channels of DCR layers (see Section~\ref{sec:recnet:arch}) and by increasing its value we approximate wider architectures. We experimented with the RecNet($e$,8,16,32,10,10,10) architecture and the results are presented at Table~\ref{table:CRC}(b). One can observe that higher values of the expansion hyper-parameter $e$ lead to both improved performance and an increase in the number of parameters. However, the accuracy gain between $e=4$ and $e=8$ is trivial compared to the increase in the number of parameters. Therefore, we conclude that $e=4$ is a good compromise with regards to the size-accuracy trade-off of the model.

\subsubsection{Kernel Size of $\mathbf{W_x}$ and $\mathbf{W_h}$}
\label{sec:exp:crc:kernel}
The CRC layer can be still effective even one of the weight tensors $\mathbf{W_h}$, $\mathbf{W_x}$ uses $3 \times 3$ convolution filters, while the other uses $1\times 1$. By selecting $1 \times 1$ convolution for either of the tensors, we further reduce the model parameters. Assuming the RecNet(4,8,16,32,10,10,10) architecture, these implementation choices are evaluated in Table~\ref{table:CRC}(c). We can observe that the use of $1\times 1$ convolutions compresses the parameters by $5-20\%$ at the cost of a small decrease in performance. Nevertheless, we shall continue the experimental section with $3\times3$ kernels for both matrices, since they correspond to the best performing setting. 

\subsubsection{Recurrent vs Grouped Convolution}
To highlight the impact of the recurrent formulation, we also implemented a network with exactly the same parameters, but without recurrence, i.e. each layer consists of a grouped convolution with shared parameters formulated as: $\mathbf{h}_i = \sigma_i(\mathbf{x_i} \circledast (\mathbf{W_x} \circledast \mathbf{W_h}))\,\forall i \in [0, d-1].$  
Table~\ref{table:CRC}(d) contains the evaluation of such a network which is considerably worse compared to the proposed one, since the CRC layer assumes a much more complex structure and larger receptive field. 

\begin{table}[!h]
\begin{tabular}{cc}
\resizebox{.48\textwidth}{!}{
\begin{tabular}{ccc}
\toprule
Non-linearity $\sigma$ & CIFAR-10 & CIFAR-100 \\
\midrule\midrule
ReLU & 94.62 & 75.48 \\
shared BN + ReLU & 87.44 & 62.70 \\
separate BN + ReLU & 95.15 & 78.25 \\
None (linear) & 94.24 & 75.71 \\
\bottomrule
\end{tabular}
}&
\resizebox{.48\textwidth}{!}{
\begin{tabular}{lccc}
\toprule
$e$ & \#params & CIFAR-10 & CIFAR-100 \\
\midrule\midrule
$1$ & 424K & 93.38 & 71.94 \\
$2$ & 824K & 94.16 & 73.10\\
$4$ (default) & 1,769K & 95.15 & 78.25\\
$8$ & 4,239K& 95.47 & 79.33\\
\bottomrule
\end{tabular}
}\\
\vspace{.1pt}\\
(a)&(b) \\
\vspace{.1pt}\\
\resizebox{.48\textwidth}{!}{
\begin{tabular}{ccccc}
\toprule
$\mathbf{W_x}$ & $\mathbf{W_h}$ & \#params & CIFAR-10 & CIFAR-100 \\
\midrule\midrule
$3\times3$ & $1\times1$ & 1,425K & 94.71 & 77.48\\
$1\times1$ & $3\times3$ & 1,683K & 94.72 & 77.42\\
$3\times3$ & $3\times3$  & 1,769K & 95.15 & 78.25\\
\bottomrule
\end{tabular}
}&
\resizebox{.48\textwidth}{!}{
\begin{tabular}{cccc}
\toprule
 & \#params & CIFAR-10 & CIFAR-100 \\
\midrule\midrule
recurrent & 1,769K & 95.15 & 78.25\\
grouped & 1,769K & 94.65 & 76.04\\
\bottomrule
\end{tabular}
}\\
\vspace{.1pt}\\
(c)&(d)\\
\vspace{.1pt}\\
\end{tabular}
\caption{Exploration of (a) non-linearities, (b) expansion hyper-parameter $e$ and (c) kernel sizes of $\mathbf{W_x}$ and $\mathbf{W_h}$ (d) recurrent vs grouped convolution with shared parameters. All experiments use the RecNet(4, 8, 16, 32, 10, 10, 10) architecture.}
\label{table:CRC}
\end{table}

\subsection{Depth/Width of RecNet}
\label{sec:exp:dw}

Having concluded on the most efficient non-linearity, expansion parameter and kernel size for the weight tensors, we proceed with the evaluation of the hyper-parameters which control the simulated width and the depth of the CRC layers, i.e. $S_1, S_2, S_3, d_1, d_2, d_3$. Specifically, we progressively increase the intermediate feature map channels ($e\cdot S_i\cdot d_{i}$) at each pair of Recurrent modules by increasing $S_i$ ($S_1 < S_2 < S_3$) while having $d_i$ fixed ($d_1 = d_2 = d_3 = d_f$) or vice versa. The results are summarized in Table~\ref{table:depth-exploration}. The reported depth corresponds to $d_1 + d_2 + d_3$ for simplicity. However, the depth of the network is $d_1 + d_2 + d_3 + 7$, if we include the first convolution layer as well as the six Transition Blocks. All the networks included at Table \ref{table:depth-exploration} perform well and are fairly compact, requiring at most a few million parameters. Compared to the popular MobileNet architecture~\cite{Howard2017MobileNetsEC} which has an accuracy of $73.65\%$ on CIFAR-100 at 3.3M parameters, RecNet(4,8,8,8,5,10,15) has an accuracy of $73.68\%$ at 0.3M parameters and RecNet(4,8,16,32,15,15,15) has an accuracy of $79.01\%$ at 3.3M parameters.

\begin{table*}[h!]
\begin{center}
\footnotesize
\begin{tabular}{lccccc}
\toprule
Architecture ($e$, $S_1$, $S_2$, $S_3$, $d_1$, $d_2$, $d_3$) & acronym & depth & \#params & CIFAR-10 & CIFAR-100 \\
\midrule\midrule
RecNet(4, 4, 8, 16, 10, 10, 10) & RecNet-60-640 & 60 & 471K & 93.45 & 74.94\\
RecNet(4, 4, 8, 16, 15, 15, 15) & RecNet-90-960 & 90 & 863K & 94.32 & 76.31 \\
RecNet(4, 4, 8, 16, 20, 20, 20) & RecNet-120-1280 & 120 & 1,406K & 94.58 & 77.62\\
\midrule
RecNet(4, 8, 16, 32, 10, 10, 10) & RecNet-60-1280 & 60 & 1,769K & 95.15 & 78.25 \\
RecNet(4, 8, 16, 32, 15, 15, 15) & RecNet-90-1920 & 90 & 3,306K & 95.22 & 79.01 \\
RecNet(4, 8, 16, 32, 20, 20, 20) & RecNet-120-2560 & 120 & 5,444K & 95.46 & 80.31 \\
\midrule
RecNet(4, 8, 8, 8, 5, 10, 15) & RecNet-60-480 & 60 & 316K & 93.28 & 73.68 \\
RecNet(4, 8, 8, 8, 10, 15, 20) & RecNet-90-640 & 90 & 537K & 93.95 & 75.86 \\
RecNet(4, 8, 8, 8, 10, 20, 30) & RecNet-120-960 & 120 & 930K & 94.44 & 77.58 \\
\midrule
RecNet(4, 16, 16, 16, 5, 10, 15) & RecNet-60-960 & 60 & 1,137K & 94.54 & 77.91 \\
RecNet(4, 16, 16, 16, 10, 15, 20) & RecNet-90-1280 & 90 & 2,028K & 94.77 & 78.00 \\
RecNet(4, 16, 16, 16, 10, 20, 30) & RecNet-120-1920 & 120 & 3,569K & 95.47 & 79.38 \\
\bottomrule
\end{tabular}
\end{center}
\caption{Exploration of hyper-parameters $S_1$, $S_2$, $S_3$, $d_1$, $d_2$, $d_3$, which control the size of the RecNet architecture. The acronym \emph{RecNet-$d$-$w$} reports a model with a depth $d$ and a maximum width $w$.}
\label{table:depth-exploration}
\end{table*}



\subsection{Comparison to the state-of-the-art}

\begin{figure}
    \centering
    \includegraphics[width=.8\textwidth]{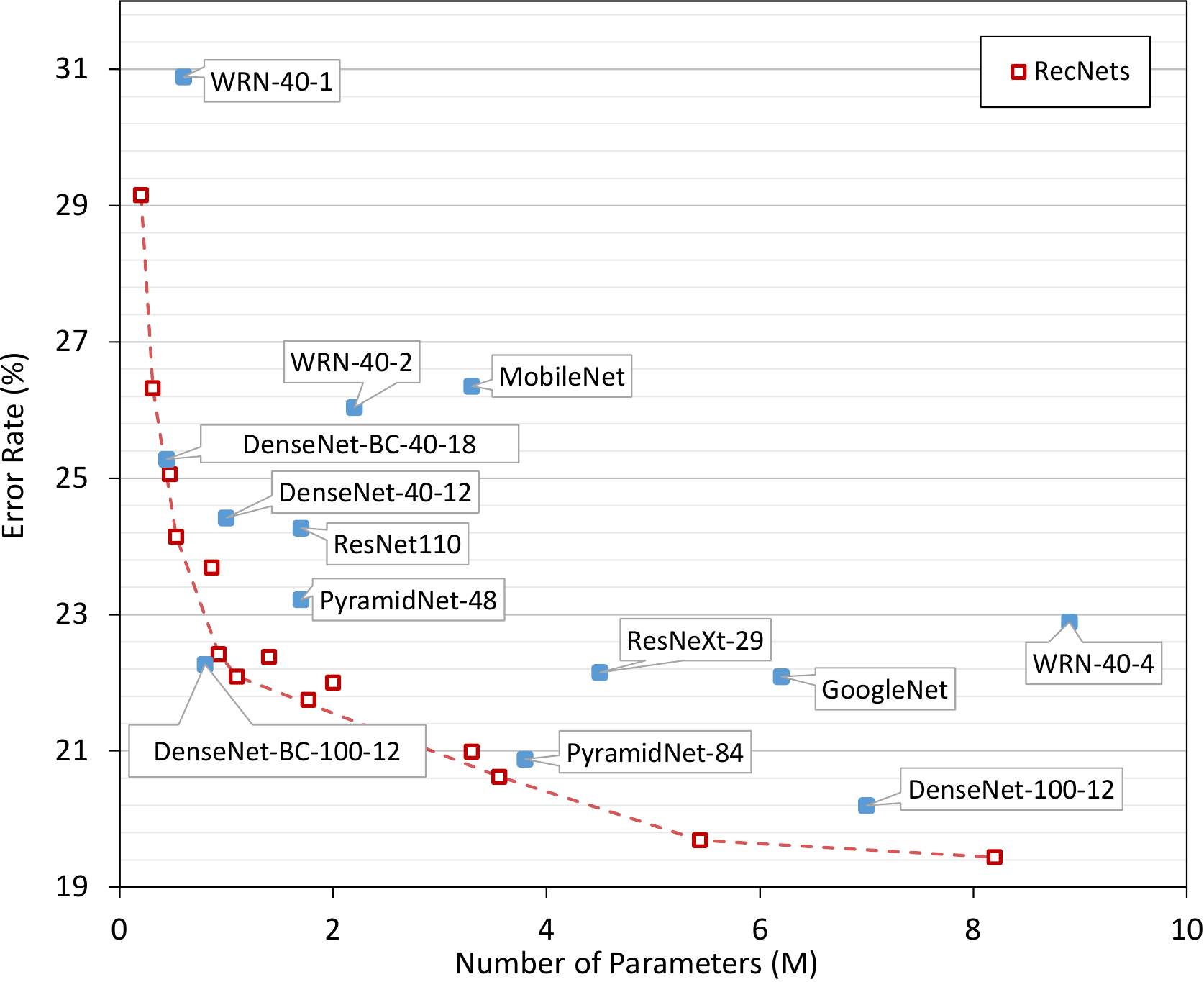}
    \caption{Comparison of state-of-the-art networks on CIFAR-100 in terms of size-accuracy trade-off.}
    \label{fig:pareto}
\end{figure}

Table~\ref{table:c100eval} shows the performance of several compact state-of-the-art models on CIFAR-100---this dataset is significantly more challenging than CIFAR-10 and, therefore, it is easier to draw meaningful conclusions about the evaluated architectures. 
We select models consisting of at most a few million parameters. 
For clarity, we only report two RecNet architectures in Table~\ref{table:c100eval}. 
Specifically, we select two extreme cases, one with only 240K parameters and one with almost 9M parameters.
Even though RecNets perform well compared to the majority of reported networks, its superiority is effectively shown in Figure~\ref{fig:pareto}.
RecNets demonstrate an improved performance-parameter trade-off compared to the majority of other networks, with the exception of DenseNets, which provide a comparable trade-off.
In fact, RecNets and DenseNets share the concept of a history preserving step,  but implement it in a different manner (hidden state of recurrent formulation in case of RecNets and densely-connected layers in case of DenseNets).
This observation hints towards a property that seems essential to constructing efficient compact networks: \emph{re-using/perplexing information at different receptive fields}.

\begin{table}[!h]
\begin{center}
\begin{tabular}{lcc}
\toprule
& \#Params & Accuracy \\
Model/Method & (Million) & (\%) \\
\midrule\midrule
GoogleNet~\cite{Szegedy_2015_CVPR} & 6.8 & 77.91 \\ %
\midrule
ResNet-110~\cite{He_2016_CVPR} & 1.7 & 75.73 \\ %
\midrule
WRN-40-1~\cite{Zagoruyko2016WideRN} & 0.6 & 69.11 \\ %
WRN-40-2~\cite{Zagoruyko2016WideRN} & 2.2 & 73.96 \\
WRN-40-4~\cite{Zagoruyko2016WideRN} & 8.9 & 77.11 \\
\midrule
MobileNet~\cite{Howard2017MobileNetsEC} & 3.3 & 73.65 \\
\midrule
ResNeXt-29,1x64d~\cite{Xie_2017_CVPR} & 4.5 & 77.85 \\ %
\midrule
PyramidNet-48~\cite{han2017deep} & 1.7 & 76.78 \\ %
PyramidNet-84~\cite{han2017deep} & 3.8 & 79.12 \\
\midrule
DenseNet-40-12~\cite{Huang_2017_CVPR} & 1.0 & 75.58 \\ %
DenseNet-100-12~\cite{Huang_2017_CVPR} & 7.0 & 79.80 \\
DenseNet-BC-40-18~\cite{Huang_2017_CVPR} & 0.4 & 74.72 \\
DenseNet-BC-100-12~\cite{Huang_2017_CVPR} & 0.8 & 77.73 \\
\midrule
RecNet-30-60 & 0.2 & 70.85 \\
RecNet-120-2880 & 8.2 & 80.56 \\
\bottomrule
\end{tabular}
\end{center}
\vspace{.1pt}
\caption{Accuracy of state-of-the-art networks ($<$10M parameters) on CIFAR-100.}
\label{table:c100eval}
\end{table}

\section{Conclusions}

In this paper we employ the recurrent logic of RNNs in order to design a novel convolutional layer, called Channel-wise Recurrent Convolutional (CRC) layer.
The input of a CRC layer is expressed as a sequence of feature maps and is processed in a recurrent fashion. 
Theoretical advantages of CRC layers are two-fold: a) they simulate wide layers with high receptive fields and b) they improve parameter efficiency due to parameter sharing and cross-layer connectivity.
Using CRC as the main building block, we present a family of compact neural network architectures, which we refer to as RecNets. 
We evaluate RecNets on the CIFAR-10 and CIFAR-100 image classification tasks and demonstrate that, for a given parameter budget, they outperform other state-of-the-art networks.

\section*{Awknoledgements}

This project has received funding from the European Union's Horizon 2020 research and innovation programme under grant agreement No 732204 (Bonseyes). This work is supported by the Swiss State Secretariat for Education, Research and Innovation (SERI) under contract number 16.0159. The opinions expressed and arguments employed herein do not necessarily reflect the official views of these funding bodies.

\bibliographystyle{unsrt}  
\bibliography{recnet}
\end{document}